\documentclass[conference]{IEEEtran}
\IEEEoverridecommandlockouts
\usepackage{cite}
\usepackage{amsmath,amssymb,amsfonts}
\usepackage{algorithmic}
\usepackage{graphicx}
\usepackage{textcomp}
\usepackage{xcolor}
\usepackage[colorlinks=true, linkcolor=black, citecolor=black, urlcolor=black]{hyperref}
\usepackage{booktabs}
\usepackage{multirow}
\usepackage{tikz}
\usetikzlibrary{arrows,positioning,calc}
\def\BibTeX{{\rm B\kern-.05em{\sc i\kern-.025em b}\kern-.08em
    T\kern-.1667em\lower.7ex\hbox{E}\kern-.125emX}}

\usepackage{caption}
\usepackage{subfloat}
\usepackage{subcaption}
\usepackage{textcomp}
\usepackage{xcolor}
\usepackage{booktabs,siunitx,tabularx, stfloats}
\usepackage[acronym, shortcuts]{glossaries}
\glsdisablehyper  
\newacronym{gai}{GenAI}{generative AI}
\newacronym{dl}{DL}{Deep Learning}
\newacronym{roc}{ROC}{Receiver Operating Characteristic}
\newacronym{ba}{BA}{Balanced Accuracy}
\newacronym{nn}{NN}{Neural Network}
\newacronym{cnn}{CNN}{Convolutional Neural Network}
\newacronym{mse}{MSE}{Mean Squared Error}
\newacronym{tp}{TP}{True Positive}
\newacronym{tn}{TN}{True Negative}
\newacronym{fp}{FP}{False Positive}
\newacronym{fn}{FN}{False Negative}
\newacronym{relu}{ReLU}{Rectified Linear Unit}
\newacronym{ml}{ML}{Machine Learning}
\newacronym{rms}{RMS}{Root Mean Square}
\newacronym{fpr}{FPR}{False Positive Rate}
\newacronym{mm}{3DMM}{3D Morphable Model}
\newacronym{pca}{PCA}{Principal Component Analysis}
\newacronym{3ddfav3}{3DDFA-V3}{3D Dense Face Alignment Version 3}
\newacronym{sgd}{SGD}{Stochastic Gradient Descent}
\newacronym{BA}{BA}{Balanced Accuracy}
\newacronym{BApoi}{$\mathrm{BA}_{\mathrm{poi}}$}{Balanced Accuracy}
\newacronym{AUCpoi}{$\mathrm{AUC}_{\mathrm{poi}}$}{AUC}
\newacronym{mlp}{MLP}{Multi-Layer Perceptron}
\newacronym{vit}{ViT}{Vision Transformer}
\newacronym{cls}{CLS}{classification}
\newacronym{clip}{CLIP}{Contrastive Language Image Pretraining}
\newacronym{poi}{POI}{Person-of-Interest}
\newacronym{ar}{AR}{Augmented Reality}
\newacronym{vr}{VR}{Virtual Reality}
\newacronym{mae}{MAE}{Masked Autoencoder}
\newacronym{ssl}{SSL}{Self-Supervised Learning}
\newacronym{vgg}{VGG}{Visual Geometry Group}
\newacronym{ntx}{NT-Xent}{Normalized Temperature-scaled Cross-Entropy}
\newacronym{auc}{AUC}{Area Under the Curve}
\newacronym{hq}{HQ}{High Quality}
\newacronym{lq}{LQ}{Low Quality}
\newacronym{bamax}{$\mathrm{BA}$}{Balanced Accuracy}
\newacronym{ba0}{$\textrm{B-ACC}_{@ \textrm{thr}=0}$}{Balanced Accuracy at threshold 0}
\newacronym{tpr0}{$\textrm{TPR}_{@ \textrm{thr}=0}$}{True Positive Rate at threshold 0}
\newacronym{tprtheta}{$\textrm{TPR}(\theta)$}{True Positive Rate at threshold $\theta$}
\newacronym{fpr0}{$\textrm{FPR}_{@ \textrm{thr}=0}$}{False Positive Rate at threshold 0}
\newacronym{lpip}{LPIPS}{Learned Perceptual Image Patch Similarity}

\newcommand{\method}{MOTIF}

\newcommand{\Dref}{\mathcal{D}_{\mathrm{ref}}}   
\newcommand{\Wset}{\mathcal{W}}                  
\newcommand{\Bank}{\mathcal{R}}                  
\newcommand{\Query}{\mathcal{Q}}                 
\newcommand{\Mouth}{\mathcal{M}}                 

\begin{document}

\title{\method: Person-of-Interest Deepfake Detection \\Beyond 3DMM Coefficients
\thanks{This work was supported by the FOSTERER project, funded by the Italian Ministry of Education, University, and Research within the PRIN 2022 program.
This work was partially supported by the European Union - Next Generation EU under the Italian National Recovery and Resilience Plan (NRRP), Mission 4, Component 2, Investment 1.3, CUP D43C22003080001, partnership on ``Telecommunications of the Future'' (PE00000001 - program ``RESTART'') and by the Investment 1.3, CUP D43C22003050001, partnership on ``SEcurity and RIghts in the CyberSpace'' (PE00000014 - program ``FF4ALL-SERICS'').}}

\author{Giovanni Affatato, Sara Mandelli,  Paolo Bestagini, Stefano Tubaro\\
Dipartimento di Elettronica, Informazione e Bioingegneria, Politecnico di Milano, 20133 Milan, Italy.\\
\small{Corresponding Author: giovanni.affatato@polimi.it}}

\maketitle

\begin{abstract}
Video deepfakes targeting a specific individual, the \gls{poi}, are the most harmful ones, and, since a public figure is abundantly recorded, a detector can be built from genuine footage of that individual.
Such detectors commonly describe a subject through a \gls{mm} and adopt its coefficients as a whole, so which part of that description carries the signal has never been measured.
We dissect it, holding the encoder, the training corpus and the enrollment protocol fixed and varying only what the encoder observes.
The groups of coefficients prove largely redundant, since the shape block alone recovers almost all the accuracy of the full vector, and their temporal evolution contributes a real but bounded amount.
We further show that the dense surface the same fit returns, which these detectors discard, carries identity information that the coefficients do not, and that it helps precisely where they are weakest.
We assemble the best configuration into \method, a visual-only detector trained on real videos only, with no manipulated video and no \gls{poi}-specific data.
It improves on both state-of-the-art \gls{poi} detectors in every dataset and manipulation of our benchmark and at two quality levels.
Our experimental code will be released at \href{https://github.com/polimi-ispl/MOTIF}{polimi-ispl/MOTIF}.
\end{abstract}

\begin{IEEEkeywords}
person-of-interest deepfake detection, 3D morphable models, facial geometry, contrastive learning.
\end{IEEEkeywords}

\glsresetall

\section{Introduction}
\label{sec:intro}

Among video deepfakes, the most harmful are those that target one specific individual, referred to as the \gls{poi}, such as a politician, a celebrity or a corporate executive, since such forgeries enable disinformation and fraud.
Generative AI keeps making such forgeries easier to produce and harder to recognize~\cite{koutlisVideoDeepfakeDetection2026}.
However, a \gls{poi} is a public figure, so genuine recordings of that subject are abundant, and a detector can be built around them.
Such detectors describe how the subject looks~\cite{dongProtectingCelebritiesDeepFake2022} or how the subject moves while speaking~\cite{agarwalProtectingWorldLeaders}.

Some of these detectors are trained on the specific \gls{poi} they protect, which requires a large amount of footage of that subject and yields a model that does not transfer to anyone else.
We address the complementary approach, which learns a general representation of identity from many subjects and enrolls the \gls{poi} only at inference, from a set of pristine reference videos, so that neither footage of the \gls{poi} nor any manipulated video is required during training~\cite{cozzolinoIDRevealIdentityawareDeepFake2021, cozzolinoAudioVisualPersonofInterestDeepFake2023, petmezasVideoDeepfake2025, salvi2025phoneme}.

Most detectors of this kind describe a subject through a \gls{mm}~\cite{cozzolinoIDRevealIdentityawareDeepFake2021, petmezasVideoDeepfake2025, shioharaExposeAnyonePersonalized2026}, a statistical model of face geometry that represents a face by a short vector of coefficients weighting a basis of aligned three-dimensional face scans~\cite{egger3DMorphableFace2020}.
Separate groups of coefficients account for the facial shape, which belongs to the subject and is stable across recordings, for the expression and for the rigid head pose, and the same coefficients decode back into a dense surface of vertices covering the face.
The appeal is twofold: the description is compact, and it keeps no pixels and hence neither the appearance of the subject nor the traces left by a generator.

These works, however, adopt the representation wholesale: a fixed vector of shape, expression and pose coefficients is fed to a temporal encoder, and the resulting system is evaluated end to end.
Their ablations vary the encoder and the objective, but not the description.
ID-Reveal compares three metric losses and measures the effect of its adversarial branch, which it introduces so that the encoder depends on behavior rather than on visual information alone~\cite{cozzolinoIDRevealIdentityawareDeepFake2021}, while Petmezas et al.\ vary the number of recurrent and attention layers and credit the result with a better modeling of the temporal dynamics of facial expressions~\cite{petmezasVideoDeepfake2025}.
In both cases the claim concerns the representation, while the experiments concern the model built upon it.
We therefore examine the description itself, leaving the encoder aside, and explore three questions whose answers would inform anyone building on a \gls{mm}: (i) which groups of coefficients provide an effective signal for detection, and whether they are complementary or redundant; (ii) whether the temporal evolution of the coefficients adds anything over a time-invariant summary; and (iii) whether the dense surface the model reconstructs, discarded once the coefficients are extracted, carries identity information that the coefficients do not.

\begin{figure}[t]
\centering
\includegraphics[width=\columnwidth]{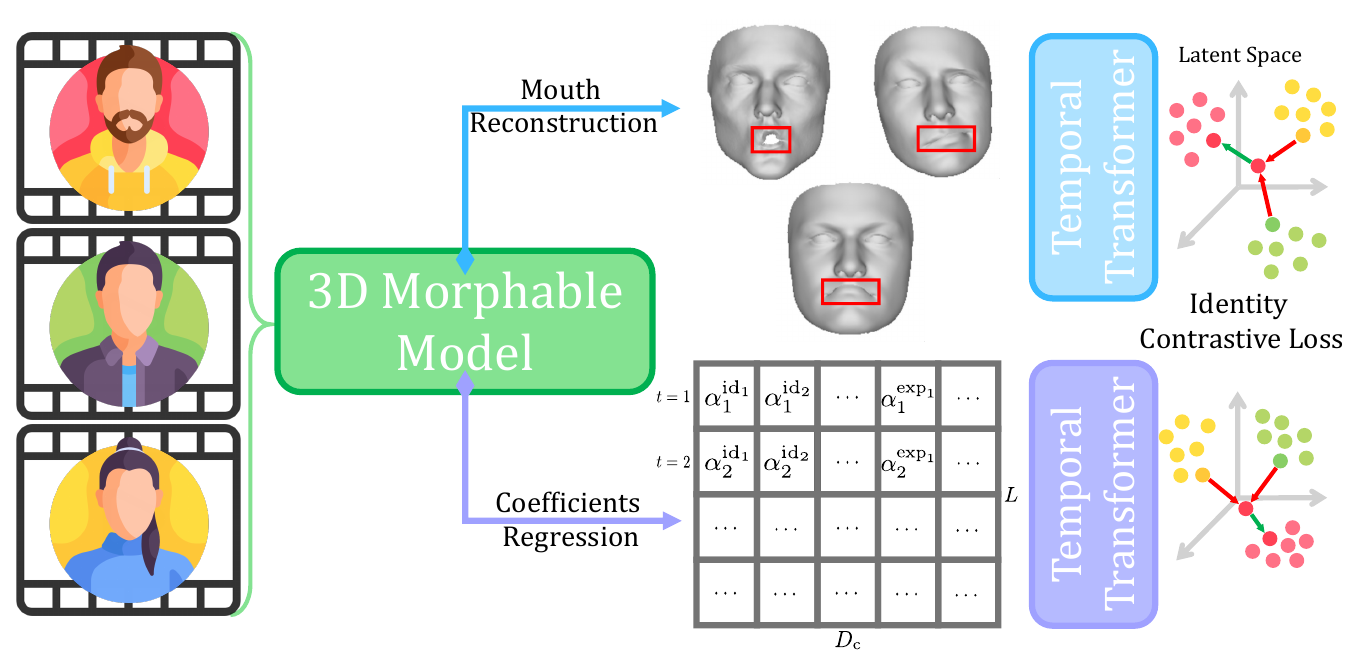}
\caption{\method\ training overview.}
\label{fig:pipeline}
\end{figure}

To answer them we build \method{} (MOrphable model for Temporal Identity Forensics), a visual-only \gls{poi} deepfake detector that serves first as a testbed and then, in its best configuration, as our proposed method.
Its encoder is a temporal transformer trained by an identity-contrastive objective~\cite{chenSimpleFrameworkContrastive2020} on real videos of many subjects, with no manipulated video and no footage of the target \gls{poi}, as sketched in Fig.~\ref{fig:pipeline}.
We retrain it from scratch for every representation we compare, holding the corpus, the encoder and the enrollment protocol fixed, so that a difference in performance is attributable to the representation alone.
We vary the groups of coefficients and their temporal treatment in a factorial design, pairing each sequence with a time-invariant summary, so that the temporal evolution is isolated from the capacity of the input.
We also decode the coefficients back into geometry~\cite{wang3DFaceReconstruction2024a} and describe the vertices of the mouth as a separate branch, evaluated alone and fused with the global one.
Our contributions are:
\begin{itemize}
  \item We present a controlled study of the \gls{mm} description for reference-based \gls{poi} deepfake detection.
  \item We introduce a use of the representation that coefficient-based detectors leave unexploited, namely the signal carried by the 3D reconstruction of the mouth.
  \item We assemble the best configuration into \method{} and compare it with six state-of-the-art detectors on four datasets spanning face swap, reenactment and lip-sync, at two levels of video quality.
\end{itemize}

\section{Proposed Methodology}
\label{sec:method}

\subsection{POI Deepfake Detection}
\label{sec:method_problem}

Each target subject, the \gls{poi}, is characterized by a collection of pristine reference videos $\Dref = \{V_1, \dots, V_{N_{\mathrm{ref}}}\}$ that genuinely depict the individual.
Given a test video $V$ claimed to portray the same \gls{poi}, our goal is to determine whether $V$ is authentic or manipulated, assigning a label $y \in \{\textrm{real}, \textrm{fake}\}$ on the basis of $\mathcal{D}_{\textrm{ref}}$.
In line with Cozzolino et al.~\cite{cozzolinoIDRevealIdentityawareDeepFake2021,cozzolinoAudioVisualPersonofInterestDeepFake2023}, we frame detection as the thresholding of a subject-similarity score.
Let $E$ denote an encoder mapping a video to a set of subject descriptors. A test video is scored according to how well its descriptors $E(V)$ align with those extracted from the reference set
via a similarity function $S(V, \mathcal{D}_{\textrm{ref}}) \in \mathbb{R}$.

We study this problem under two constraints.
First, we operate in an \emph{open-set} regime, meaning that the forgery techniques employed to generate deepfakes are not known at training time.
Second, we require detection to be \emph{\gls{poi}-agnostic}: we do not use \gls{poi}-specific data at training time, so that the encoder is not tailored to any particular identity.

\subsection{Extraction of Facial Representations}
\label{sec:repr}

\subsubsection{Global descriptor extraction}
Let a video be a sequence of $T$ frames.
We process the $t$-th frame with a \gls{mm} face reconstruction model~\cite{wang3DFaceReconstruction2024a}, which regresses compact sets of coefficients describing the observed face: the \emph{shape} coefficients $\boldsymbol{\alpha}^{\mathrm{id}}_t \in \mathbb{R}^{D_{\mathrm{id}}}$, which encode the neutral facial geometry of the subject, the \emph{expression} coefficients $\boldsymbol{\alpha}^{\mathrm{exp}}_t \in \mathbb{R}^{D_{\mathrm{exp}}}$, which encode the facial gesture being performed, the \emph{head pose} $\boldsymbol{\rho}_t \in \mathbb{R}^{3}$ and the \emph{translation} $\mathbf{d}_t \in \mathbb{R}^{3}$, which describe the rigid motion of the head with respect to the camera.
We concatenate the four retained coefficient sets into a single per-frame vector
\begin{equation}
  \mathbf{c}_t = \big[\,\boldsymbol{\alpha}^{\mathrm{id}}_t \,;\,
  \boldsymbol{\alpha}^{\mathrm{exp}}_t \,;\, \boldsymbol{\rho}_t \,;\,
  \mathbf{d}_t \,\big] \in \mathbb{R}^{D_{\mathrm{c}}},
  \label{eq:coeff}
\end{equation}
with $D_{\mathrm{c}} = D_{\mathrm{id}} + D_{\mathrm{exp}} + 3 + 3$, which we normalize using the 
mean and standard deviation estimated on real training videos.

Both in training and in test, we process a video $V$ in windows of $L$ consecutive frames with $50\%$ overlap.
The \emph{global} descriptor, which observes the whole face, is the window starting at the $t$-th frame,
\begin{equation}
  \mathbf{X}^{\mathrm{global}}_t =
  \big[\,{\mathbf{c}}_t, \dots, {\mathbf{c}}_{t+L-1}\,\big]^{\!\top}
  \in \mathbb{R}^{L \times D_{\mathrm{c}}} .
  \label{eq:window}
\end{equation}
\subsubsection{Mouth descriptor extraction}
The coefficients summarize the whole face into a single vector, whereas the same fit also returns a dense surface, whose vertices place the description on specific parts of the face.
We therefore extract a second descriptor from the lip vertices alone.

Using \glspl{mm}, we reconstruct the 3D geometry of the face from each frame and retain only the lip vertices.
We refer the reader to~\cite{wang3DFaceReconstruction2024a} for the details of the reconstruction.
Let $\Mouth$ be the index set of the lip vertices of the face model extracted through the \gls{mm} and let $D_{\mathrm{v}}$, the number of vertex coordinates, be equal to $3\lvert \Mouth \rvert$.
We then subtract the temporal mean of the lip vertices over the analysis window, which removes the subject's neutral lip shape and retains only how the lips move, independently of what they look like at rest.
We define the mouth descriptor as $\mathbf{X}^{\mathrm{mouth}}_t \in \mathbb{R}^{L \times D_{\mathrm{v}}}$.

In general, we denote by $\Wset_b(V)$ the set of the window descriptors of branch $b$ extracted from a video $V$, so that $\Wset_{\mathrm{global}}(V)$ collects the global face descriptors, while $\Wset_{\mathrm{mouth}}(V)$ the mouth ones.
We denote by $L \times D_b$ the input dimension of branch $b$.

\subsection{Encoder and Training Objective}
\label{sec:encoder}

\subsubsection{Temporal transformer}
Both branches (global and mouth) employ the same architecture, which we define as the map $E_b : \mathbb{R}^{L \times D_b} \rightarrow \mathbb{S}^{C-1}$.
In practice, a linear layer tokenizes each frame into a $C$-dimensional token, a learnable \gls{cls} token is prepended, learnable positional embeddings are added and a stack of transformer blocks processes the sequence.
We define $E_b(\mathbf{X})$ as the $\ell_2$-normalized \gls{cls} embedding of the last block.
Notice that attention operates over time only, so that the model can relate distant frames of the same window.
The two instances of $E_b$ are trained separately and share no weights, and they differ only in the input dimension $D_b$ of the tokenizer.

\subsubsection{Identity-contrastive objective}
We train each encoder with a single identity-contrastive objective.
Let $\mathcal{B}$ be the set of window indices in a batch and let $\ell(a)$ be the identity label of window $a$.
We define the set of positive samples of an anchor $a \in \mathcal{B}$ as all the other windows of the same subject, i.e., $\mathcal{P}(a) = \{\, j \in \mathcal{B} \setminus \{a\} : \ell(j) = \ell(a) \,\}$, while every window of a different subject is a negative sample.
Let $\mathcal{K}$ denote the set of selected encoder depths, and let $\mathbf{z}^{(k)}_a \in \mathbb{S}^{C-1}$ be the \gls{cls} embedding of window $a$ read at layer $k$.
We adopt the \gls{ntx} loss~\cite{chenSimpleFrameworkContrastive2020}, which we write for an anchor $a$ and a depth $k$ as
\begin{equation}
  \resizebox{.9\columnwidth}{!}{$\displaystyle
  \mathcal{L}^{(k,a)} = -\frac{1}{\lvert \mathcal{P}(a) \rvert}
  \sum_{j \in \mathcal{P}(a)} \log
  \frac{\exp\!\big(\langle \mathbf{z}^{(k)}_a, \mathbf{z}^{(k)}_j \rangle /
  \eta\big)}
  {\sum\limits_{m \in \mathcal{B} \setminus \{a\}}
  \exp\!\big(\langle \mathbf{z}^{(k)}_a, \mathbf{z}^{(k)}_m \rangle / \eta\big)}
  $}
  \label{eq:ntxent}
\end{equation}
here $\eta$ is the temperature parameter and $\langle \cdot , \cdot \rangle$ denotes the inner product.
The total objective aggregates over all the anchors and the selected depths with equal weight.
In doing so, we make the intermediate blocks, which present features with an increasing order of abstraction, identity-discriminative as well.

We populate $\mathcal{P}(a)$ with windows drawn from different videos of the same subject, so that most positive pairs are cross-video and the objective cannot be satisfied by recording-specific cues.
Moreover, we populate $\mathcal{B}$ by balancing the identities, so that no subject dominates the contrastive comparison.
We also use a Multi-Similarity Miner~\cite{wangMultiSimilarityLossGeneral2019}, which keeps only the most informative comparisons.

Furthermore, the contrastive pass operates on a temporally masked window in which contiguous spans of frames are removed, so that the two views of a positive pair are not trivially matchable.
Masking is applied at training time only.

\subsection{Reference-Set Matching and Score Fusion}
\label{sec:matching}

\subsubsection{Reference-set matching}
To enroll the \gls{poi}, we compute for each branch $b$ a set of reference embeddings $\Bank_b$, one for every reference window,
\begin{equation}
  \Bank_b = \big\{\, E_b(\mathbf{X}) \;:\;
  \mathbf{X} \in \Wset_b(V_i),\; V_i \in \Dref \,\big\}.
  \label{eq:bank}
\end{equation}
A query video $V$ yields in the same way $\Query_b(V) = \{ E_b(\mathbf{X}) : \mathbf{X} \in \Wset_b(V) \}$.

The reference embeddings of a subject are usually not isotropic: they spread along the directions in which that subject varies across genuine videos, such as recording conditions and spoken content, and stay narrow along the directions in which that subject is stable.
A plain cosine weights every direction equally, so it rewards agreement inside a spread that any video of the subject already covers, and it dilutes a disagreement along the narrow directions, which is where identity actually lives.
We therefore rescale each direction by the inverse of its reference standard deviation, so that a query earns similarity only where the subject is consistent.

We define the covariance of $\Bank_b$ to be $\boldsymbol{\Sigma}_b$. 
We define the per-\gls{poi} whitening map as
\begin{equation}
  \omega_b(\mathbf{z}) =
  \frac{\boldsymbol{\Sigma}_b^{-1/2}\,\mathbf{z}}
       {\big\lVert \boldsymbol{\Sigma}_b^{-1/2}\,\mathbf{z} \big\rVert_2} ,
  \label{eq:whiten}
\end{equation}
where $\lVert \cdot \rVert_2$ denotes the Euclidean norm. We score the query as the maximum cosine similarity in the whitened space,
\begin{equation}
  S_b(V; \Bank_b) = \max_{\mathbf{r} \in \Bank_b}\;
  \max_{\mathbf{q} \in \Query_b(V)}
  \big\langle \omega_b(\mathbf{r}), \omega_b(\mathbf{q}) \big\rangle ,
  \label{eq:score}
\end{equation}
where $\mathbf{r}$ and $\mathbf{q}$ denote a reference and a query embedding.
\subsubsection{Score fusion}
The two scores of~\eqref{eq:score} live on per-subject and per-branch scales, so before combining them we calibrate them on the reference set.
Specifically, we score each reference video in a leave-one-out setting, i.e., every $i$-th reference video is tested against the set $\Bank_b^{(-i)}$ of the reference embeddings of $\Dref \setminus \{V_i\}$. We denote by $\mu_b$ and $\sigma_b$ the mean and the standard deviation of the scores $\{S_b(V_i; \Bank_b^{(-i)})\}_{i=1}^{N_{\mathrm{ref}}}$ so obtained.
A generic query video (not part of the reference set) is then expressed in units of the genuine variability of that subject,
\begin{equation}
  z_b(V) = \frac{S_b(V; \Bank_b) - \mu_b}{\sigma_b} .
  \label{eq:znorm}
\end{equation}

This operation places the two branches on a common scale, so that they can be combined with a straightforward arithmetic mean, where each of them can raise or lower the verdict and a query is authentic only if it is consistent with the reference set both in the description of the whole face and in that of the mouth.

\section{Experimental Setup}
\label{sec:setup}

\subsection{Datasets}
\label{sec:datasets}

\subsubsection{Training data}
We train on real videos of VoxCeleb2~\cite{chungVoxCeleb2DeepSpeaker2018}, an in-the-wild audio-visual corpus of interview footage.
It comprises $5{,}994$ identities, $150{,}480$ videos and $1.13$M clips.
The pristine videos of FakeAVCeleb are themselves VoxCeleb2 clips, so we remove from the training list the $500$ identities used to build that dataset~\cite{khalidFakeAVCelebNovelAudioVideo2022}.
In this way, none of the identities we train on appears among the evaluation \glspl{poi}.

\subsubsection{Evaluation data}
We evaluate on four public deepfake datasets, namely DF-TIMIT~\cite{korshunovDeepFakesNewThreat2018}, FakeAVCeleb~\cite{khalidFakeAVCelebNovelAudioVideo2022}, DeepSpeak~\cite{barringtonDeepSpeakDataset2025} and KoDF~\cite{kwonKoDFLargescaleKorean2021}.
Together, they span all the main manipulation families that a detector has to face, namely identity swapping, full-face reenactment or synthesis, identity-preserving lip-sync, and their combinations.

We group the manipulations in Table~\ref{tab:catmap} according to what they do to the two quantities that a \gls{poi} detector can observe, i.e., facial \emph{appearance} and facial \emph{motion}: (i) \emph{Face-swap} (FS): the face of the target is replaced by that of a source identity, so appearance is altered globally while the driving performance remains that of the target; (ii) \emph{Reenactment} (RE): identity and appearance are preserved, or synthesized as an avatar, but the motion is imported from a driving actor, i.e., the complement of the previous case; (iii) \emph{Lip-sync} (LS): identity, appearance and head motion are all preserved, and only the mouth is resynthesized to match a target audio track; (iv) \emph{Face-swap\,+\,lip-sync} (FS+LS): a face swap that has then been lip-synced, so both quantities are altered at once.

\begin{table}[t]
\centering
\caption{Generation methods of each evaluation dataset, grouped into the four
canonical categories used in the per-category results.}
\label{tab:catmap}
\resizebox{\columnwidth}{!}{
\begin{tabular}{p{1.5cm} l p{4.3cm}}
\toprule
Category & Dataset & Generation method \\
\midrule
\multirow{4}{=}{\textit{Face-swap} (FS)}
            & DF-TIMIT    & autoencoder-based swap \\[2pt]
            & FakeAVCeleb & FaceSwap, FSGAN \\[2pt]
            & DeepSpeak   & FaceFusion, INSwapper, SimSwap \\[2pt]
            & KoDF        & FaceSwap, DeepFaceLab, FSGAN \\
\midrule
\textit{Face-swap\,+\,lip-sync} (FS+LS)
            & FakeAVCeleb & FaceSwap and FSGAN followed by Wav2Lip~\cite{prajwalLipSyncExpert2020} \\
\midrule
\multirow{2}{=}{\textit{Reenactment} (RE)}
            & DeepSpeak   & LivePortrait, HelloMeme, Memo \\[2pt]
            & KoDF        & FOMM \\
\midrule
\multirow{3}{=}{\textit{Lip-sync} (LS)}
            & FakeAVCeleb & Wav2Lip~\cite{prajwalLipSyncExpert2020} \\[2pt]
            & DeepSpeak   & Diff2Lip, LatentSync \\[2pt]
            & KoDF        & Wav2Lip~\cite{prajwalLipSyncExpert2020} \\
\bottomrule
\end{tabular}
}
\end{table}

\subsection{Evaluation Metrics}
\label{sec:metrics}

We assess performance through the \gls{auc} of the \gls{roc} curve, which measures the quality of the score ranking regardless of the operating threshold, and the \gls{ba}, which we take as the maximum of the balanced accuracy over the decision threshold.
We compute both per \gls{poi}, i.e., we score the real videos of a subject against its fakes, and we average the result over the subjects of a dataset.
This is consistent with the \gls{poi} scenario, in which the system targets a \textit{specific subject} and a small set of subject-specific samples is available to calibrate the threshold, so that the reported values measure the discriminative capability of the method after adaptation to the target \gls{poi}.
When we report a single manipulation category, we restrict the fakes of each subject to that category, we leave its real videos unchanged and we exclude the subjects that have no fake of it.
The aggregate values instead use all the fakes of a subject, regardless of the manipulation.

\subsection{State-of-the-Art Baselines}
\label{sec:baselines}

We compare \method\ against publicly available general and \gls{poi}-specific deepfake detectors. 
Among general deepfake detectors, we consider RealForensics~\cite{haliassosLeveragingRealTalking2022b}, LipForensics~\cite{haliassosLipsDontLie2021a} and FTCN~\cite{zhengExploringTemporalCoherence2021}.
We select them because they release pretrained weights that we can evaluate directly under our cross-dataset protocol, their common training corpus being FaceForensics++~\cite{rosslerFaceForensicsLearningDetect2019a}, which shares no material with any dataset we test on.
We further include the model of Seferbekov~\cite{seferbekovSelimsefDfdc_deepfake_challenge2026}, which won the DFDC challenge and was trained on the corresponding dataset.

Regarding \gls{poi}-specific detectors, we select two recently proposed and publicly available methods, namely POI-Forensics~\cite{cozzolinoAudioVisualPersonofInterestDeepFake2023} and ID-Reveal~\cite{cozzolinoIDRevealIdentityawareDeepFake2021}.
Both are trained on VoxCeleb2~\cite{chungVoxCeleb2DeepSpeaker2018}, i.e., the corpus we train on as well, which makes them the closest available comparison to \method.
POI-Forensics is a multimodal framework that analyzes audio and video jointly: we evaluate it in its video-only setting so that the comparison matches our visual-only setup.
ID-Reveal is visual-only by design and is the prior method closest to ours, since it also describes a subject through \gls{mm} coefficients.

\subsection{Architecture and Training Details}
\label{sec:details}

\subsubsection{Input}
We extract the per-frame coefficients with the 3DDFA-V3 model~\cite{wang3DFaceReconstruction2024a}, of which we keep $D_{\mathrm{id}}=80$ shape coefficients, $D_{\mathrm{exp}}=64$ expression coefficients, $3$ head-pose angles and $3$ translation components, for a total of $D_{\mathrm{c}}=150$ dimensions per frame.
We process each video in windows of $L=75$ frames, i.e., $3$~sec at $25$~fps, with $50\%$ overlap.
The mouth branch reconstructs the $\lvert\Mouth\rvert=968$ lip vertices of the face model, i.e., $D_{\mathrm{v}}=2{,}904$ dimensions per frame, from which we remove the per-window temporal mean before applying the same per-dimension scaling.

\subsubsection{Encoders}
Both branches employ the same temporal transformer, with $12$ blocks, hidden size $C=192$ and $3$ attention heads, preceded by a linear per-frame tokenizer and by learnable positional embeddings.

\subsubsection{Training}
We train with the \gls{ntx} objective at temperature $\eta=0.1$, applied with equal weight to the \gls{cls} embedding read at the depths $\mathcal{K}=\{3,6,9,12\}$, and we mine the pairs with a Multi-Similarity Miner ($\epsilon=0.1$).
Batches are drawn by the identity-balanced sampler as $8$ subjects $\times$ $4$ videos $\times$ $2$ windows, and the training-time view of a window is masked by contiguous spans of $15$ frames covering half of it.
We optimize with AdamW, at learning rate $10^{-4}$ and weight decay $0.1$, under a cosine schedule with $10$ warm-up epochs.
We select the deployed checkpoint as the epoch of lowest contrastive loss on the held-out VoxCeleb2 test identities, which comprise $118$ identities and $942$ videos and are disjoint from the training ones by the split of the dataset.

\section{Results}
\label{sec:results}

\begin{table*}[t]\centering
\caption{\gls{auc} / \gls{ba} (\%) per dataset and manipulation category, considering static (S) coefficients or dynamic (D) ones. Best result per column in bold. }
\label{tab:dynamic_retrained}
\resizebox{\textwidth}{!}{
\begin{tabular}{lccccccccccccc}
\toprule
 & \multicolumn{4}{c}{FakeAVCeleb} & \multicolumn{4}{c}{DeepSpeak} & \multicolumn{4}{c}{KoDF} & \multicolumn{1}{c}{Overall}\\
\cmidrule(lr){2-5}\cmidrule(lr){6-9}\cmidrule(lr){10-13}\cmidrule(lr){14-14}
Input & FS & FS+LS & LS & All & FS & RE & LS & All & FS & RE & LS & All & Mean\\
\midrule
Shape (S) & 91.6/92.2 & 93.1/93.5 & \textbf{60.7/78.3} & 85.3/87.0 & 98.5/98.4 & 87.6/90.9 & 86.7/91.5 & 89.7/91.3 & 97.6/97.4 & 85.7/86.8 & 93.0/93.5 & 93.1/92.0 & 89.4/90.1\\
Expression (S) & 89.2/90.4 & 91.1/91.7 & 57.0/75.7 & 82.9/85.1 & 98.3/98.5 & 88.2/91.4 & 87.2/91.3 & 90.0/91.2 & \textbf{97.8/97.7} & 85.8/87.0 & 94.0/94.2 & 93.5/92.3 & 88.8/89.6\\
Shape\,+\,expr. (S) & 90.5/91.6 & 91.8/92.9 & 59.0/78.3 & 83.9/86.3 & 98.3/98.3 & 87.9/90.8 & 86.3/91.6 & 89.6/91.1 & 97.6/97.6 & 86.6/87.5 & 96.2/95.5 & 93.7/92.6 & 89.0/90.0\\
\midrule
Shape (D) & \textbf{92.8/93.4} & 93.3/94.1 & 59.2/77.7 & 85.6/87.6 & \textbf{99.7/99.6} & \textbf{90.7/92.6} & \textbf{89.3/92.5} & \textbf{92.2/92.5} & 96.9/96.8 & 86.5/87.4 & 95.5/95.4 & 93.4/92.4 & 90.4/90.8\\
Expression (D) & 90.2/91.0 & 91.7/92.8 & 57.9/77.0 & 83.8/85.6 & 99.6/99.7 & 90.2/92.8 & 88.9/92.6 & 91.8/92.8 & 97.5/97.3 & 86.4/87.3 & 95.5/95.9 & 93.6/92.5 & 89.7/90.3\\
Shape\,+\,expr. (D) & 92.6/93.7 & \textbf{95.0/95.7} & 58.3/77.1 & \textbf{86.2/87.9} & 99.5/99.6 & 90.5/92.9 & 88.0/91.9 & 91.6/92.3 & 97.5/97.4 & \textbf{86.8/86.7} & \textbf{96.4/95.8} & \textbf{94.0/92.6} & \textbf{90.6/90.9}\\
\bottomrule
\end{tabular}
}
\end{table*}

\subsection{Dissecting the Coefficient Vector}
\label{subsec:results_dynamic}

Prior work adopts the \gls{mm} coefficients as a whole, so we first ask which of their groups provides an effective signal for detection, and whether the groups are complementary or redundant.
We then ask whether the evolution of the coefficients over time carries information that a time-invariant description of the same subject does not.

We design the experiment as a factorial ablation over the coefficient vector of~\eqref{eq:coeff}, which we split into a shape block $\mathbf{s}_t = \boldsymbol{\alpha}^{\mathrm{id}}_t$ and an expression block $\mathbf{e}_t = [\,\boldsymbol{\alpha}^{\mathrm{exp}}_t \,;\, \boldsymbol{\rho}_t \,;\, \mathbf{d}_t\,]$, which collects the expression, the pose and the translation.
We train the encoder with $\mathbf{s}_t$, with $\mathbf{e}_t$ or with $\mathbf{c}_t$, and we read each of them in a \emph{dynamic} variant, which keeps the per-frame trajectory over the window, and in a \emph{static} variant, which repeats the average of the block over the clip, obtaining the six configurations of Table~\ref{tab:dynamic_retrained}, from which we omit DF-TIMIT because every configuration achieves $100\%$ on it.

The three inputs turn out to be largely interchangeable.
The six configurations span less than two points of overall mean \gls{auc}, and the three dynamic ones lie within $0.9$ points of each other.
Reading the shape block alone attains $90.4\%$ against the $90.6\%$ of the full vector, so the expression, the pose and the translation are worth $0.2$ points once the shape is read over time, whereas discarding the shape block costs $0.9$.

Reading the coefficients dynamically is the better choice for every block, and it is worth $1.0$ points of overall mean \gls{auc} on the shape block, $0.9$ on the expression block and $1.6$ on their combination.
The gain is largest on reenactment, where the motion of a driving actor is precisely the manipulated quantity.

Both results have a common explanation in the front end.
The shape of a face does not change while a subject speaks, so its coefficients should in principle be constant in time and their clip average should be the best available estimate of that constant, yet reading the block over time is worth a point, so its variation is not estimation noise.
Since we regress the coefficients independently on each frame, nothing keeps the shape coefficients of a clip consistent, and since the shape and the expression bases deform the same mesh, a single frame does not separate them.
We therefore conjecture that the shape block absorbs a share of the expression dynamics during the per-frame fit, which would explain both why it gains from being read over time and why the two blocks stay so close to each other.

Along both axes, then, the description is less differentiated than its common use would suggest: the groups of coefficients largely duplicate each other, and their temporal evolution contributes real but bounded information.
Most of the accuracy of the full vector is already available from the shape block alone.
We nonetheless adopt the full coefficient vector $\mathbf{c}_t$, read dynamically, for the remainder of this article, since it is the best configuration on average.

\begin{table*}[t]\centering
\caption{\gls{auc} / \gls{ba} (\%) per dataset and manipulation category, considering the global branch only (G-only), the mouth branch only (M-only) or their fusion.
Best result per column is highlighted in bold.}
\label{tab:mouth}
\resizebox{\textwidth}{!}{
\begin{tabular}{lcccccccccccccc}
\toprule
 & \multicolumn{1}{c}{DF-TIMIT} & \multicolumn{4}{c}{FakeAVCeleb} & \multicolumn{4}{c}{DeepSpeak} & \multicolumn{4}{c}{KoDF} & \multicolumn{1}{c}{Overall}\\
\cmidrule(lr){2-2}\cmidrule(lr){3-6}\cmidrule(lr){7-10}\cmidrule(lr){11-14}\cmidrule(lr){15-15}
Config & FS & FS & FS+LS & LS & All & FS & RE & LS & All & FS & RE & LS & All & Mean\\
\midrule
G-only & \textbf{100.0/100.0} & \textbf{92.6/93.7} & \textbf{95.0/95.7} & 58.3/77.1 & 86.2/87.9 & \textbf{99.5/99.6} & \textbf{90.5/92.9} & \textbf{88.0/91.9} & \textbf{91.6/92.3} & \textbf{97.5/97.4} & 86.8/86.7 & 96.4/95.8 & 94.0/92.6 & 92.9/93.2\\
M-only & 92.1/92.8 & 76.9/82.2 & 82.4/85.6 & \textbf{68.2/79.0} & 77.6/81.5 & 88.3/91.5 & 81.0/84.9 & 84.3/88.3 & 84.0/86.3 & 85.0/86.5 & 81.7/84.6 & 90.8/91.4 & 85.2/85.6 & 84.7/86.6\\
Fusion & \textbf{100.0/100.0} & 91.0/92.3 & 93.3/94.0 & 67.4/79.7 & \textbf{86.8/87.7} & 99.1/99.3 & 88.5/91.6 & 87.5/91.8 & 90.5/91.6 & 97.1/97.1 & \textbf{90.4/90.6} & \textbf{98.0/97.4} & \textbf{95.3/94.0} & \textbf{93.1/93.3}\\
\bottomrule
\end{tabular}
}
\end{table*}

\subsection{Exploiting the Reconstructed Surface}
\label{subsec:results_mouth}

The coefficients are not the only description a \gls{mm} provides, since the same fit also returns a dense surface, so we ask whether that surface carries identity information that the coefficients do not.
Table~\ref{tab:mouth} reports the global and the mouth branch read alone and fused by the mean of their per-\gls{poi} calibrated scores.

Read alone, the mouth branch attains an overall mean \gls{auc} of $84.7\%$ against the $92.9\%$ of the global branch, so a few hundred vertices of the reconstructed lips already identify a subject almost as well as the whole coefficient vector.
What matters for our question, however, is not how far the surface goes on its own but whether it adds anything to the coefficients, and it does: fusing the two branches improves on the global one by $9.1$ points on the lip-sync manipulation of FakeAVCeleb, by far the weakest column of the table, and by $3.6$ points on the reenactment of KoDF.

These two columns have little in common as manipulations, which already suggests that what the surface supplies does not depend on the manipulation being confined to the mouth.
The lip-sync columns of KoDF and FakeAVCeleb make the point directly, since they come from the same generator (Table~\ref{tab:catmap}) and yet the global branch attains $96.4\%$ on the first and $58.3\%$ on the second, so that the fusion adds $1.6$ points instead of $9.1$: the surface adds most where the coefficients achieve least.

Across the remaining columns the fusion never loses more than $2.0$ points, and it attains the best overall mean of the three configurations at $93.1\%$, so the information it contributes comes at a bounded cost.
We adopt the fusion for the remainder of this article.

\begin{table*}[t]\centering
\caption{State-of-the-art comparison per manipulation (\gls{auc} / \gls{ba}, \%). \textbf{Bold}: best \gls{poi}-specific method; \textit{italic}: best general deepfake detector.}
\label{tab:sota_comparison}
\resizebox{\textwidth}{!}{
\begin{tabular}{lcccccccccccccc}
\toprule
 & \multicolumn{1}{c}{DF-TIMIT} & \multicolumn{4}{c}{FakeAVCeleb} & \multicolumn{4}{c}{DeepSpeak} & \multicolumn{4}{c}{KoDF} & \multicolumn{1}{c}{Overall}\\
\cmidrule(lr){2-2}\cmidrule(lr){3-6}\cmidrule(lr){7-10}\cmidrule(lr){11-14}\cmidrule(lr){15-15}
Method & FS & FS & FS+LS & LS & All & FS & RE & LS & All & FS & RE & LS & All & Mean\\
\midrule
\multicolumn{15}{c}{\textit{High quality}}\\
\midrule
RealForensics~\cite{haliassosLeveragingRealTalking2022b} & \textit{100.0/100.0} & \textit{98.2/98.8} & 95.7/94.3 & 75.7/78.9 & 86.9/84.9 & 86.5/90.0 & 84.9/84.0 & 86.1/86.7 & 85.5/83.5 & 87.1/85.5 & \textit{94.5/91.4} & 99.2/98.1 & 93.1/89.9 & 91.4/89.6\\
LipForensics~\cite{haliassosLipsDontLie2021a} & 97.9/98.1 & 96.5/96.2 & 96.1/94.4 & \textit{95.7/94.3} & \textit{96.3/93.8} & 84.1/89.5 & \textit{93.7/93.2} & \textit{97.8/97.4} & \textit{93.6/92.6} & 88.7/88.8 & 91.2/88.5 & \textit{99.6/99.1} & 93.1/90.8 & \textit{95.2/93.8}\\
FTCN~\cite{zhengExploringTemporalCoherence2021} & \textit{100.0/100.0} & 80.5/84.9 & 86.8/87.4 & 80.8/84.8 & 82.6/82.7 & \textit{88.7/92.2} & 73.4/80.5 & 83.2/87.0 & 79.4/83.3 & 90.3/89.8 & 92.0/88.5 & 98.2/97.5 & \textit{93.5/90.8} & 88.9/89.2\\
Seferbekov~\cite{seferbekovSelimsefDfdc_deepfake_challenge2026} & 90.0/92.5 & 94.3/93.5 & \textit{98.0/97.1} & 81.1/88.1 & 89.2/89.0 & 57.2/74.8 & 59.3/69.6 & 92.9/92.9 & 70.0/75.7 & \textit{92.9/94.4} & 81.7/81.0 & 98.4/97.9 & 90.9/89.6 & 85.0/86.7\\
\cmidrule(lr){1-15}
POI-Forensics~\cite{cozzolinoAudioVisualPersonofInterestDeepFake2023} & 89.6/93.3 & 81.6/83.8 & 76.5/80.5 & 52.8/71.5 & 66.7/73.1 & 94.9/96.8 & 65.1/73.8 & 43.4/59.9 & 62.0/68.4 & 95.4/94.6 & 71.1/71.2 & 80.9/78.2 & 84.5/81.8 & 75.7/79.2\\
ID-Reveal~\cite{cozzolinoIDRevealIdentityawareDeepFake2021} & 97.6/97.4 & 68.9/75.7 & 76.5/79.1 & 59.7/72.0 & 67.1/72.1 & 93.9/95.6 & 72.2/77.0 & 64.4/73.9 & 72.9/75.8 & 93.6/91.2 & 67.0/67.9 & 87.2/82.8 & 83.0/80.1 & 80.1/81.4\\
Ours & \textbf{100.0/100.0} & \textbf{90.8/90.9} & \textbf{92.8/92.2} & \textbf{63.7/75.5} & \textbf{78.8/79.4} & \textbf{99.4/99.5} & \textbf{88.0/90.0} & \textbf{87.3/90.3} & \textbf{89.4/89.6} & \textbf{96.9/96.1} & \textbf{89.7/86.2} & \textbf{97.2/95.0} & \textbf{94.8/91.1} & \textbf{90.8/90.0}\\
\midrule
\multicolumn{15}{c}{\textit{Low quality}}\\
\midrule
RealForensics~\cite{haliassosLeveragingRealTalking2022b} & 77.0/84.4 & \textit{82.2/83.5} & 54.5/65.5 & 34.6/57.9 & 51.3/62.1 & \textit{61.1/74.7} & \textit{60.6/68.4} & \textit{83.5/83.6} & \textit{68.8/72.0} & 74.4/74.8 & 82.0/79.6 & 93.3/88.6 & 81.4/78.8 & 69.6/74.3\\
LipForensics~\cite{haliassosLipsDontLie2021a} & 73.3/82.2 & 80.4/82.7 & 59.0/67.3 & \textit{59.1/67.1} & 64.2/68.0 & 60.8/74.4 & 48.1/61.1 & 74.0/77.3 & 59.1/65.9 & 79.3/80.0 & \textit{83.2/80.8} & \textit{96.7/94.0} & \textit{85.1/82.8} & \textit{70.4/74.7}\\
FTCN~\cite{zhengExploringTemporalCoherence2021} & \textit{88.5/91.0} & 34.2/59.4 & 24.4/55.1 & 33.2/59.6 & 30.7/56.4 & 58.6/72.6 & 59.5/68.0 & 58.1/67.6 & 58.7/65.8 & 56.8/62.2 & 66.3/67.2 & 79.3/76.6 & 64.3/65.7 & 60.6/69.7\\
Seferbekov~\cite{seferbekovSelimsefDfdc_deepfake_challenge2026} & 55.3/72.8 & 79.2/80.4 & \textit{82.9/83.1} & 45.3/65.1 & \textit{64.3/70.1} & 53.0/71.3 & 49.7/61.6 & 60.8/68.4 & 54.0/62.2 & \textit{82.8/83.8} & 53.6/60.3 & 81.9/80.2 & 73.1/74.0 & 61.7/69.8\\
\cmidrule(lr){1-15}
POI-Forensics~\cite{cozzolinoAudioVisualPersonofInterestDeepFake2023} & 88.4/92.8 & 81.3/83.3 & 76.3/80.1 & 53.5/71.0 & 66.9/72.8 & 95.1/96.8 & 61.5/70.2 & 40.9/58.2 & 59.2/65.8 & 95.3/94.5 & 70.0/70.3 & 79.7/77.0 & 83.8/81.2 & 74.6/78.1\\
ID-Reveal~\cite{cozzolinoIDRevealIdentityawareDeepFake2021} & 97.8/98.1 & 66.5/73.6 & 71.7/75.7 & 55.9/68.8 & 63.2/69.3 & 93.4/95.2 & 66.8/72.5 & 55.0/66.3 & 66.8/70.3 & 92.4/89.8 & 63.6/65.5 & 82.7/79.1 & 80.3/77.9 & 77.0/78.9\\
Ours & \textbf{100.0/100.0} & \textbf{86.7/87.0} & \textbf{88.9/88.1} & \textbf{60.5/71.3} & \textbf{75.2/75.8} & \textbf{98.8/98.9} & \textbf{76.5/79.3} & \textbf{71.5/76.7} & \textbf{78.0/78.0} & \textbf{96.4/95.7} & \textbf{82.8/79.8} & \textbf{94.9/91.1} & \textbf{91.6/87.6} & \textbf{86.2/85.3}\\
\bottomrule
\end{tabular}
}
\end{table*}

\subsection{Comparison with the State of the Art}
\label{subsec:results_sota}

Table~\ref{tab:sota_comparison} positions \method{} against the two families of baselines on all four datasets and at two quality levels, where we refer to every original dataset as \gls{hq} and we define its \gls{lq} version as an H.264 recompression at CRF~$40$, which emulates the lossy re-encoding applied by video-sharing platforms.

In the \gls{hq} block our method is competitive with the general deepfake detectors, since LipForensics~\cite{haliassosLipsDontLie2021a} attains the highest mean \gls{auc} and \method{} trails it by $4.4$ points, while among the \gls{poi}-specific methods it leads by $10.7$ points over ID-Reveal and by $15.1$ over POI-Forensics.
This comparison is not symmetric in terms of supervision, since the general detectors are trained on manipulated videos, whereas our method never observes one.

The picture changes under compression, where \method{} becomes the best method overall, ahead of the second best by $9.2$ points.
It gives up $4.6$ points of mean \gls{auc} between the two blocks, against an average of $24.6$ for the general detectors, which is consistent with what the two families measure: the latter rely on low-level synthesis traces that a recompression at CRF~$40$ largely destroys, whereas the facial description survives it.
The \gls{poi}-specific baselines are comparably stable, losing $1.1$ and $3.1$ points, yet they start from a considerably lower level.

\section{Conclusions}
\label{sec:conclusions}

We have taken apart the \gls{mm} description that reference-based \gls{poi} detectors adopt as a whole, and measured what each of its parts contributes to the detection.
Its groups of coefficients turn out to be largely redundant, and their temporal evolution to be worth a real but bounded amount, while the dense surface that the same fit returns, and that these detectors discard, carries information the coefficients do not.
Assembling the best configuration into \method{} improves on both state-of-the-art \gls{poi} detectors in every cell of our benchmark and at both quality levels.
A \gls{mm} fitted consistently over a clip, rather than frame by frame, would separate identity from behavior more sharply, and facial regions other than the mouth remain to be explored.

\bibliographystyle{IEEEtran}
\bibliography{biblio, biblio2}

\end{document}